\documentclass{article}

\usepackage{arxiv}

\usepackage[utf8]{inputenc} % allow utf-8 input
\usepackage[T1]{fontenc}    % use 8-bit T1 fonts
\usepackage{hyperref}       % hyperlinks
\usepackage{url}            % simple URL typesetting
\usepackage{booktabs}       % professional-quality tables
\usepackage{amsfonts}       % blackboard math symbols
\usepackage{nicefrac}       % compact symbols for 1/2, etc.
\usepackage{microtype}      % microtypography
\usepackage{lipsum}		% Can be removed after putting your text content
\usepackage{graphicx}
\usepackage{multirow}
\usepackage{doi}
\usepackage{siunitx}
\usepackage{ragged2e}

\usepackage{algorithm}
\usepackage{algpseudocode}
\usepackage{amsmath}

\algnewcommand\algorithmicinput{\textbf{Input:}}
\algnewcommand\Input{\item[\algorithmicinput]}
\algnewcommand\algorithmicoutput{\textbf{Output:}}
\algnewcommand\Output{\item[\algorithmicoutput]}
\algnewcommand\algorithmicdefine{\textbf{Define:}}
\algnewcommand\Define{\item[\algorithmicdefine]}

\usepackage{amsmath}
\usepackage{amssymb} % For \mathbb{R} and \mathcal{H}
\usepackage{braket}

\usepackage{authblk}
\usepackage{amsmath}
\usepackage{amssymb}
\usepackage{subcaption}
\usepackage{float}
\usepackage{array}

\title{QuXAI: Explainers for Hybrid Quantum Machine Learning Models}

\author[1,a]{Saikat Barua}
\author[2]{Mostafizur Rahman}
\author[3]{Shehenaz Khaled}
\author[4]{Md Jafor Sadek}
\author[5]{Rafiul Islam}
\author[6]{Dr. Shahnewaz Siddique}

\affil[1]{North South University, Dhaka, \texttt{saikat.barua@northsouth.edu}}
\affil[a]{SpontAlign, Dhaka}
\affil[2]{North South University, Dhaka, \texttt{mostafizur.rahman10@northsouth.edu}}
\affil[3]{North South University, Dhaka, \texttt{shehenaz.khaled@northsouth.edu}}
\affil[4]{North South University, Dhaka, \texttt{jafor.sadek@northsouth.edu}}
\affil[5]{North South University, Dhaka, \texttt{rafiul.islam19@northsouth.edu}}
\affil[6]{Associate Professor, North South University, Dhaka, \texttt{shahnewaz.siddique@northsouth.edu}}

\date{}

\renewcommand{\shorttitle}{QuXAI}

\hypersetup{
  pdftitle={QuXAI: Explainers for Hybrid Quantum Machine Learning Models},
  pdfsubject={A framework and methodology for eXplainable Artificial Intelligence (XAI) applied to hybrid quantum-classical machine learning models, detailing the QuXAI Python package and the MEDLEY explainer.},
  pdfauthor={Saikat Barua, Mostafizur Rahman, Shehenaz Khaled, Md Jafor Sadek, Rafiul Islam, Dr. Shahnewaz Siddique},
  pdfkeywords={Quantum Machine Learning, Explainable AI, Hybrid Quantum-Classical Models, Interpretability, Feature Importance, Global Explanations}
}

\begin{document}
\maketitle

\begin{abstract}

The emergence of hybrid quantum-classical machine learning (HQML) models opens new horizons of computational intelligence but their fundamental complexity frequently leads to black box behavior that undermines transparency and reliability in their application. Although XAI for quantum systems still in its infancy, a major research gap is evident in robust global and local explainability approaches that are designed for HQML architectures that employ quantized feature encoding followed by classical learning. The gap is the focus of this work, which introduces QuXAI, an framework based upon Q-MEDLEY, an explainer for explaining feature importance in these hybrid systems. Our model entails the creation of HQML models incorporating quantum feature maps, the use of Q-MEDLEY, which combines feature based inferences, preserving the quantum transformation stage and visualizing the resulting attributions. Our result shows that Q-MEDLEY delineates influential classical aspects in HQML models, as well as separates their noise, and competes well against established XAI techniques in classical validation settings. Ablation studies more significantly expose the virtues of the composite structure used in Q-MEDLEY. The implications of this work are critically important, as it provides a route to improve the interpretability and reliability of HQML models, thus promoting greater confidence and being able to engage in safer and more responsible use of quantum-enhanced AI technology.

Our code and experiments are open-sourced at: \url{https://github.com/GitsSaikat/QuXAI}

\end{abstract}

% keywords can be removed

\section{Introduction}

The emergence of Quantum Machine Learning (QML), as a field promises novel computational paradigms and enhanced learning capabilities \cite{Mardirosian2019QuantumEnhanced, Macaluso2024Supervised}. Near-term applications often utilize hybrid quantum-classical (HQML) algorithms, in which quantum processors manage to solve such tasks as data representation or kernel evaluation, with classical computers focusing on optimization and control \cite{Weigold2021Patterns, Mangini2021Models}. While this pragmatic approach accelerates the exploration of quantum-enhanced learning, it often inherits the "black box" nature common to complex classical models, obscuring the internal decision-making processes.  This aspect of QML systems lack of transparency presents a great barrier to wider acceptance and trust in the QML systems. Thus, building strong eXplainable AI (XAI) techniques customized for these hybrid architectures must be considered. Such explainability is not only important to debug and validate models but also to establish user trust, regulatory compliance and finally open up the potential of QML for scientific discovery and real-world problem-solving by ensuring that their operational logic is scrutable.

The pressing need for interpretability in QML systems has led to research on transferring classical XAI principles and generation of new quantum-specific explanation mechanisms. Approaches like Shapley values have been extended to measure the influence of components in quantum circuits \cite{Heese2023Explaining, Burge2023Shapley}, and there are quantum iterations of LIME (Q-LIME) that intend to give local instance-based explanations for quantum neural networks \cite{Pira2023Interpretability}. Other explorations include learning capability of parameterized quantum circuits \cite{Heimann2022Learning} and diagnostics such as quantum neural tangent kernels \cite{Scala2025Practical}. However, while a significant fair amount of great, overarching explainability approaches exist, there is a conspicuous lack of these approaches developed specifically for the common HQML paradigm in which classical input features are transformed via a quantum feature map into a new form (e.g. amplitudes of the state vector or the kernel matrices) on which a classical learner then operates Existing tools tend to either treat the whole HQML system as a black box or aim at explaining quantum circuit in isolation of the hybrid data flow without describing the impact of initial classical attributes through the hybrid data flow. Quantum Computing is error prone, quantum error correction causes additional overhead \cite{barua2023rescued}; therefore, as QML models, including those which could be used for tasks such as enhancing the error correction, become more complicated, the necessity for the tools to understand such behavior becomes of even greater importance to make such models trusted and operational. In order to handle this, we present QuXAI, a unified scheme with a Q-MEDLEY explainer, which gives global feature importance scores by carefully examining perturbations at the classical input level and tracing their effects through quantum encoding and subsequent classical learning process.

Our methodology within the QuXAI framework encompasses three core components: (i) the construction of HQML models, wherein classical input data is encoded into quantum states using defined quantum feature maps (e.g., RX rotations for amplitude encoding, or quantum kernel functions), and the resulting quantum-derived representations are then processed by standard classical machine learning algorithms; (ii) the Q-MEDLEY explainer, which synthesizes insights from Drop-Column Importance (DCI) and Permutation Importance (PI) by perturbing original classical features and, critically, re-evaluating the quantum feature mapping stage for each perturbation before assessing the impact on the classical learner's performance, thereby respecting the hybrid data flow; and (iii) a visualization module that renders these feature importance scores as accessible bar charts. This approach is novel in its specific adaptation to the operational pipeline of feature-encoding HQML models. Unlike generic model-agnostic explainers that do not differentiate the quantum processing step, or pure quantum XAI methods that might focus on circuit parameters, Q-MEDLEY directly attributes importance to the initial classical features based on their influence throughout the entire hybrid model. This tailored design, drawing inspiration from robust classical explainability techniques \cite{barua2023kaxai}, allows for a more faithful and nuanced interpretation of feature contributions in such systems.

Our empirical studies show the practicality of the QuXAI framework. We first show that the wide variation of amplitude-encoded HQML models devised are competitive in their predictive performance to their classical counterparts, even under the condition of additional noisy and redundant attributes appending the datasets, thus validating them as legitimate objects for explanation. Whilst applied to these HQML models, the Q-MEDLEY explainer systematically produces a coherent feature importance hierarchy that is capable of discriminating between original, meaningful features from synthetic noise or redundancies in different datasets, as demonstrated for the Noisy Iris (Figure~\ref{fig:qmedley_iris}) and Noisy Wine (Figure~\ref{fig:qmedley_wine}) datasets. Moreover, rigorous validation of Q-MEDLEY in the classical ML settings, with interpretable models as ground truth, demonstrates its good performance – we achieve high Recall@3 scores (Figure~\ref{fig:recall_comparison}), and Spearman rank correlations (Figure~\ref{fig:spearman_heatmap}) which are often comparable, Ablation studies (Table~\ref{tab:qmedley_ablation_recall}) also reinforce that the composite architecture employed by Q-MEDLEY, which is a combination of an adaptive weighting approach and interaction-aware mechanisms, leads to an improvement in its robustness and accuracy of feature attribution. 

The main contributions of this paper are as follows:

\begin{itemize}
    \item The development of Q-MEDLEY, a novel global and local feature importance explainer specifically designed for hybrid quantum-classical machine learning models that utilize quantum feature encoding. 
    \item The introduction of the QuXAI framework, an integrated environment for training, evaluating, and explaining HQML models, complete with data processing and visualization capabilities along with a detailed mathematical foundation.
    \item An in-depth analysis of Q-MEDLEY’s core components, including adaptive weighting and interaction-aware permutation importance. This study highlights how each component contributes to the overall effectiveness of the explainer and provides general insights into the design principles of interpretable machine learning systems.
\end{itemize}

The rest of the article is organized as follows: Section 2 reviews related works. Section 3 describes the methodology used in this study. Section 4 presents the results obtained. Section 5 discusses how the results achieved our research objectives. Finally, Section 6 concludes with remarks on future work.

\section{Related Works}

\begin{figure}[htbp] % [H] from float package tries to place it "Here"
    \centering
    \includegraphics[width=0.6\textwidth]{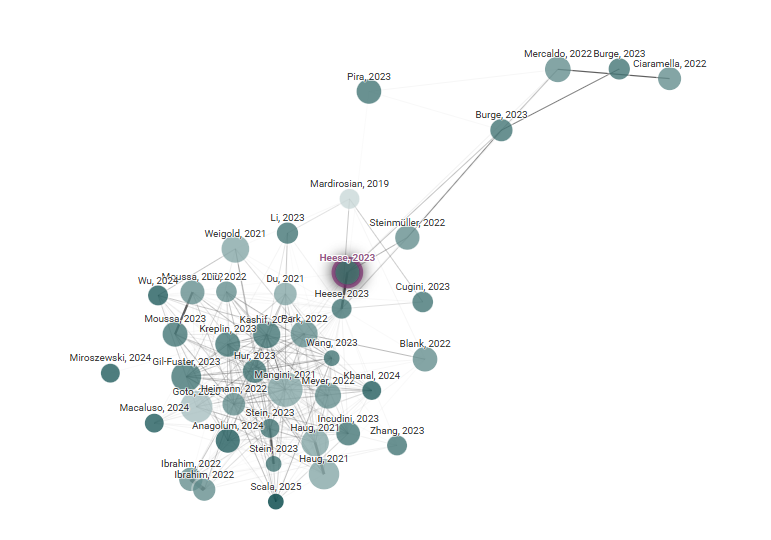} % Adjust width as needed
    \caption{Citation network map illustrating the relationships and thematic clustering of key research papers relevant to the QuXAI framework, hybrid quantum-classical models, and explainable AI in quantum machine learning. Node size represent citation counts or other metrics, and edges indicate citation links or strong thematic overlap.}
    \label{fig:citation_network}
\end{figure}

Numerous near-term QML applications use hybrid quantum-classical algorithms, wherein quantum processors are tasked with specific, frequently data representation or kernel evaluation oriented, tasks, which are executed by classical computers responsible for optimization and control \cite{Weigold2021Patterns, Mangini2021Models}. This pragmatic approach, however, receives the plight of the “black box” common for classical deep learning. And as quantum systems, built upon variational quantum circuits (VQCs) for supervised learning \cite{Mardirosian2019QuantumEnhanced}, etc., that are increasingly faced with more complex tasks, the necessity for translucence becomes more acute. Nonetheless, a straightforward transcription of eXplainable AI (XAI) methods from classical machine learning to a quantum context poses several non-trivial challenges, mainly as a result of the intrinsic probabilistic mouse on quantum measurements and the exponential scaling of quantum state-space dimensions, presenting insurmountable stumbling blocks to the contemporary implementations of XAI \cite{Steinmuller2022eXplainable}. In order to add our study to the broader context of quantum machine learning and explainability, a map of central literature is given in Figure~\ref{fig:citation_network} specifying key connections and themes.

\subsection{Shedding Light on the Quantum Black Box}

One of the important directions on explainability of Quantum Models is the use of Shapley values, a notion borrowed from cooperative game theory – to associate importance scores with various parts of a quantum computation. Scientists have shown that Shapley values are instrumental in decoupling parameterized quantum circuits, measuring the effect of single gates or groups of gates on the circuit’s behavior for particular tasks, such as classification/generative modeling \cite{Heese2023Explaining, Heese2023ExplainableQML}. This helps to explain why a specific circuit configuration works and, also, provides insight into what informs robust PQC design. Despite the problem being intractable, however, the computational load of computing Shapley values is still a concern. To solve this problem, development of specialized quantum algorithms is under way that can calculate these values with polynomial efficiency, which makes such explanations more practicable for complex QML models \cite{Burge2023Algorithm, Burge2023Shapley}. Beyond these global, model-centric explanations, methods of local, instance-specific interpretability are also comes in handy. For instance, Q-LIME, a quantum version of the classical LIME scheme, has been suggested for explaining the predictions of quantum neural networks for single data points, having provided a means of delimiting areas where predictions may be governed by the natural stochasticity of quantum measurements rather than learned aspects \cite{Pira2023Interpretability}. The applicability of such XQML techniques is investigated in various practical applications, such as improving the reliability of QML systems for mobile malware detection through revealing the characteristics determining which classification decision-taking is made \cite{Mercaldo2022MobileMalware}.

\subsection{Innovations on Architectures and Feature Representation}

The efficiency and explainability of QML systems are inherently linked with the quantum model architectures that underlies the techniques of data representation. Variational Quantum Circuits (VQCs) provide a flexible backbone for many QML algorithms including applications from supervised learning \cite{Mardirosian2019QuantumEnhanced} to quantum reinforcement learning, where improved optimized action decoding within policy gradient frameworks achieve performance advantages on both simulators and early quantum hardware \cite{Meyer2022Policy}. The design of the PQC ansatz is also an active area of research, where AI inspired approaches such as reducing width QNNs are developed to alleviate issues such as barren plateaus and increased trainability, \cite{Stein2023ReducingWidth, Stein2023AIInspired}. Architecture aspects also reach distributed quantum computing where new entanglement connection methods between multi-layer hardware-efficient ansätze can be searched for tasks such as distributed Variational Quantum Eigensolvers \cite{Zhang2023Entanglement}. The systematic comparison of various architectures of PQC regarding metric of learning capability, is vital to develop design guidelines, and understand their expressivity \cite{Heimann2022Learning}.

Classical-quantum interface is a fundamental assumption of the hybrid QML that is implemented through the quantum feature engineering. Neural quantum encoding (NQE) uses deep learning to optimize the representation of classical data in quantum states for high classification accuracy \cite{Hur2023Neural}, which is also theoretically proven for universal approximation in quantum feature maps \cite{Goto2020Universal}. Quantum kernel methods develop with compact circuits for binary classifiers \cite{Blank2022Compact} and scalable variational algorithms such as VQASVM \cite{Park2022Variational} showed satisfactory results. Quantum Hartley Transform for complex distributions, among other primitives, is beneficial in generative modeling \cite{Wu2024Hartley}. Analytic theories of wide QNN dynamics \cite{Liu2022Analytic} are used to achieve richer understanding that is applied to QML in high-energy physics \cite{Cugini2023Comparing} and to cybersecurity \cite{Ciaramella2022Introducing}.

\subsection{Addressing Trainability, Noise, and Generalization}

Major challenges are inherent in the path to utilize the complete capacity of QML, especially with regard to model trainability, ability to resist noise, and ability to generalize from a small sample set. Finite-sampling noise, a given concern in estimating expectation values from quantum circuits, can derail training and degradate performance. Approaches like variance regularization are being designed to suppress this noise through direct inclusion in the training system partly without the need for extra quantum circuit verifying ideas, hence increasing the convergence rate and producing better output quality \cite{Kreplin2023Reduction}. The performance of QML models is also highly sensitive to hyperparameters choices. Systematic studies that use functional ANOVA for example are absolutely necessary for determining the most important hyperparameters ( e.g., the learning rate, ansatz structure), and for designing data-driven prior beliefs to direct more effective hyperparameter optimization searches over a variety of data sets \cite{Moussa2023Hyperparameter, Moussa2022HyperparameterImportance}.

It is imperative to make Quantum Neural Network (QNN) predictions a priori so that tools such as Quantum Neural Tangent Kernel (QNTK) provide diagnostic abilities of model design and resource allocation \cite{Scala2025Practical}. Generalization in Noisy Intermediate-Scale Quantum (NISQ) QML is a major concern of interest with current surveys describing error bounds \cite{Khanal2024Generalization}. The classical generalization theories are challenged because of the possibility of the QNNs to fit random data, and it is necessary to have new perspectives \cite{GilFuster2023Understanding}. To improve trainability and generalization, approaches like reduced-domain initialization are explored \cite{Wang2023Enhanced}. The choice between cost function design (global vs. local) has a huge effect on training dynamics\cite{Kashif2023Impact}, while efficient shot estimation is critical for quantum kernel methods \cite{Miroszewski2024Estimating}.

\subsection{Scaling, Optimization, and Resource Efficiency}

Turning postulated QML promises into advantage, the quantum community requires collective efforts in algorithm scaling, optimization and resource efficiency as well. Frameworks such as Élivágar of the Quantum Circuit Search (QCS) type give a promising way for automating the design of high quality and noise-resistant PQCs by intelligently traversing complicated architectural spaces, taking hardware constraints into account \cite{Anagolum2024Elivagar}. To address bigger problems, quantum processor can solve, distributed quantum computing paradigms can be further explored. Examples of QUDIO are schemes that attempt to speed up variational quantum algorithms by distributing learning problems across several local quantum processors coordinated by a classical server\cite{Du2021Accelerating}. Another limit is managing big datasets which is a standard need in real-world machine learning. Examples of innovations include using randomized measurements to estimate quantum kernels, so that the quantum computational time, scales linearly, in contrast to quadratically, and thus allow for QML with larger inputs, \cite{Haug2021LargeScale, Haug2021Randomized}.

Limitations on available resources of current and near-term quantum devices are another driver of exploration of model compression and efficiency methods. Adaptation of ensemble learning methods – such as bagging and AdaBoost, which are believed to been employed within classical ML – are being explored for QNNs. Not only do these provide a mechanism to build more powerful/robust models from many, possibly weaker/smaller, quantum models, but they can also ameliorate the effects of hardware noise and also happen to save a certain amount of quantum resources \cite{Incudini2023Resource}. Finally, bridging the gap between abstract algorithms of quantum computing and concrete hardware implementation, we need optimization at the lowest possible levels. Quantum devices can be customized through the pulse-level control of quantum gates, and entanglers in PQCs. This pulse-level optimization has been demonstrated to drastically cut down state preparation times, preserve or improve circuit expressibility and facilitate VQAs trainability on real quantum hardware in a much more efficient and effective manner, opening doors to more efficient QML use \cite{Ibrahim2022Evaluation, Ibrahim2022PulseLevel}.

The joint breakthroughs on these interrelated fields, from the underlying XAI ideas for the quantum systems to revolutionary model architectures, strong training schemes and hardware aware optimizations, are moving the frontier of quantum machine learning towards effective practice of the real world applications.

\section{Methodology}

\begin{figure}[htbp] 
    \centering
    \includegraphics[width=0.3\textwidth]{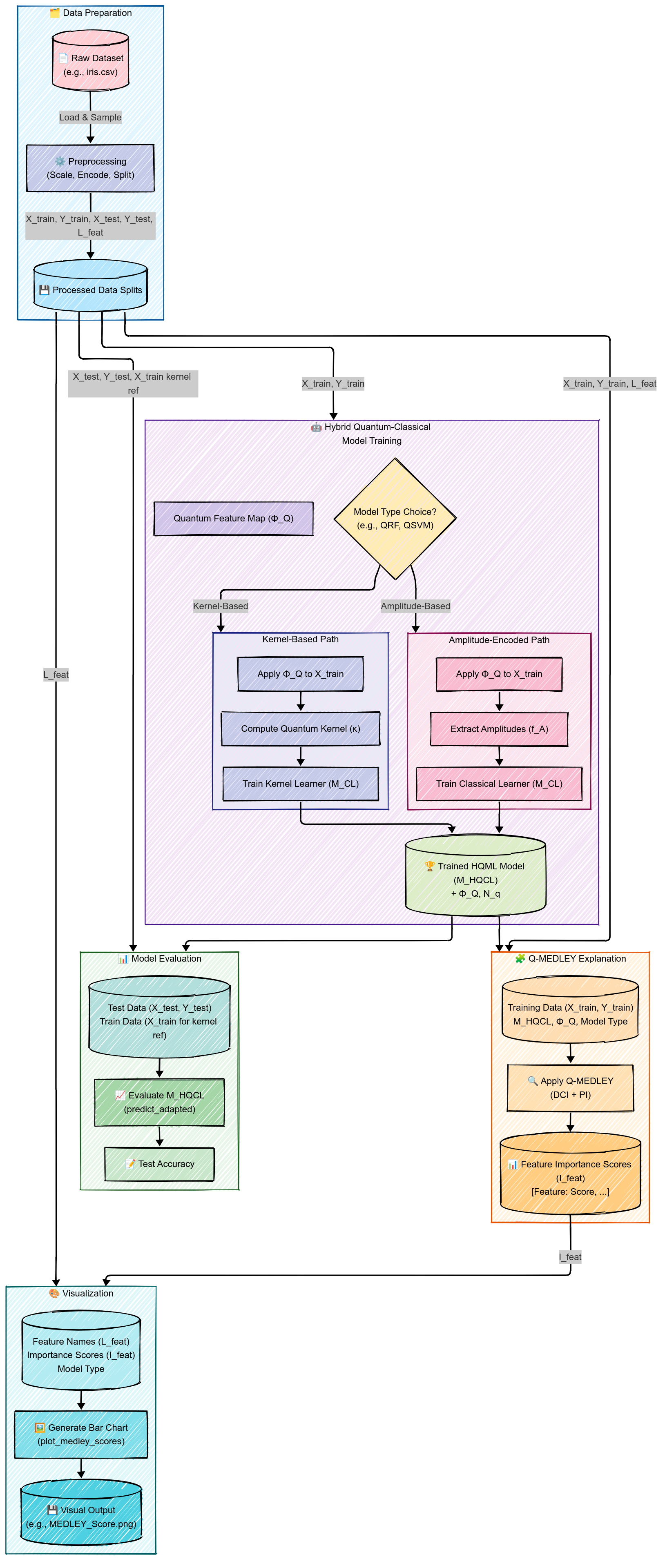}
    \caption{Overall workflow of the QuXAI framework, illustrating the sequential stages from data preparation, through hybrid quantum-classical model (HQML) training and evaluation, to feature importance explanation using Q-MEDLEY and subsequent visualization of the results. The training stage depicts distinct paths for amplitude-encoded and kernel-based HQML models. The complete pipeline is detailed in Algorithm~\ref{alg:quxai_framework_concise}.}
    \label{fig:quxai_workflow}
\end{figure}

Our work introduces QuXAI, a comprehensive framework intended to assist with both the process of training and explanation of hybrid quantum-classical machine learning (HQML) models. There are three major components of the framework: the model building of HQML models using quantum feature maps, a custom Q-MEDLEY explainer tailored for hybrid architectures and a visualization module to make sense of the outputs of the explanations. In our approach, the HQML models use quantum circuits to carry data that can either be the encoding of classical features to quantum amplitudes or construction of quantum kernels which are further managed by classical algorithms. Q-MEDLEY, an explainer that we created, combines insights from Drop-Column and Permutation Importance paradigms in order to provide strong global feature importance metrics, accounting for quantum encoding effects very carefully. Finally, the framework incorporates a way of visualizing the importance scores into a simple set of bar charts that facilitates easy evaluation of feature significance. The algorithm \ref{alg:quxai_framework_concise} guides the system through data transformation, HQML training, performance evaluation, Q-MEDLEY integration for feature interpretation and visualization of results, each step of the way ensuring that this hybrid quantum machine learning workflow is robust and interpretable. The workflow of the QuXAI framework is illustrated in ~\ref{fig:quxai_workflow}.

\subsection{Q-MEDLEY}

The Q-MEDLEY explainer introduces a robust methodology for quantifying global feature importance in hybrid quantum-classical machine learning (HQML) architectures. These HQML models, denoted as $M_{HQCL}(x) = M_{CL}(f_Q(\Phi_Q(x)))$, typically involve a two-stage process: an initial quantum feature map $\Phi_Q: \mathbb{R}^D \rightarrow \mathcal{H}$ that transforms classical $D$-dimensional input vectors $x \in X$ into quantum states $\ket{\psi(x)} \in \mathcal{H}$, followed by a classical learning algorithm $M_{CL}$ that operates on a classical representation $f_Q(\ket{\psi(x)})$ derived from these quantum states (e.g., state vector amplitudes or kernel evaluations). Q-MEDLEY's design, inspired by the MEDLEY explainer concept proposed for classical models in KAXAI \cite{barua2023kaxai}, is predicated on the principle that a synthesis of multiple, distinct feature attribution techniques can yield more stable and comprehensive insights than any single method alone. It achieves this by integrating two established perturbation-based approaches: Drop-Column Importance (DCI) and Permutation Importance (PI). This combined approach shares the goal of providing feature-level insights, similar to efforts adapting Shapley values \cite{Heese2023Explaining} or LIME \cite{Pira2023Interpretability} for quantum circuits, but distinguishes itself by its specific aggregation strategy and direct applicability to HQML feature vectors. The core rationale is that DCI assesses a feature's necessity by observing performance degradation upon its complete removal (approximated by neutralization to $x_j=0$), while PI evaluates its predictive value by disrupting its relationship with the target variable $Y$ through random shuffling. By systematically applying these perturbations to a reference dataset $(X_{ref}, Y_{ref})$ and meticulously evaluating the HQML model's response, Q-MEDLEY aims to provide a holistic measure of each feature's contribution to the model's overall predictive accuracy, $\mathcal{A}$.

The internal procedure of Q-MEDLEY is critically adapted to the hybrid nature of $M_{HQCL}$. Given a trained model $M_{HQCL}$ and the reference data $(X_{ref}, Y_{ref})$, a baseline performance $\mathcal{A}_{base} = \mathcal{A}(M_{HQCL}(X_{ref}), Y_{ref})$ is established. To compute DCI for the $j$-th feature, $X_{ref}$ is modified to $X_{ref}^{(j,0)}$ by replacing its $j$-th column with a neutral vector (e.g., $x_j \leftarrow 0$), and the importance is given by:
\begin{equation}
I_j^{DCI} = \mathcal{A}_{base} - \mathcal{A}(M_{HQCL}(X_{ref}^{(j,0)}), Y_{ref}).
\label{eq:dci}
\end{equation}
For PI, the $j$-th column of $X_{ref}$ is randomly permuted $K$ times, creating datasets $X_{ref}^{(j, \pi_k)}$ where $k=1, \dots, K$, and the importance is:
\begin{equation}
I_j^{PI} = \mathcal{A}_{base} - \frac{1}{K}\sum_{k=1}^{K} \mathcal{A}(M_{HQCL}(X_{ref}^{(j, \pi_k)}), Y_{ref}).
\label{eq:pi}
\end{equation}
The crucial step is the evaluation of $M_{HQCL}$ on these perturbed datasets. If $M_{HQCL}$ is amplitude-based, where $f_Q(\ket{\psi(x)})$ are derived from the amplitudes (e.g., $f_Q(\ket{\psi(x)}) = [|\braket{0|\psi(x)}|^2, \dots, |\braket{2^N-1|\psi(x)}|^2]$ for an $N$-qubit system), then for any perturbed input $x'$, the quantum feature map $\Phi_Q(x')$ must be re-evaluated to obtain the new amplitudes for $M_{CL}$. If $M_{HQCL}$ is kernel-based, where $M_{CL}$ utilizes a quantum kernel $K(x_i, x_l) = \kappa(\Phi_Q(x_i), \Phi_Q(x_l)) = |\braket{\psi(x_i)|\psi(x_l)}|^2$, then evaluating on a perturbed dataset $X'_{ref}$ requires computing a new kernel matrix $K(X'_{ref}, X_{ref})$ by re-evaluating the kernel function $\kappa$ between perturbed instances and the original reference training instances. The final importance for feature $j$ is an aggregation:
\begin{equation}
I_j = \frac{1}{2}(I_j^{DCI} + I_j^{PI}).
\label{eq:medley_final}
\end{equation}

The significance of Q-MEDLEY lies in its tailored approach to providing global feature explanations for this distinct class of HQML models. Unlike generic model-agnostic explainers, Q-MEDLEY's internal prediction mechanism explicitly accounts for the quantum feature encoding stage, $x \xrightarrow{\Phi_Q} \ket{\psi(x)} \xrightarrow{f_Q} \text{classical features/kernel}$. This ensures that the impact of perturbations to the original classical features is correctly propagated through the quantum transformation $\Phi_Q$ before being assessed by the classical learner $M_{CL}$. This nuanced handling is essential for accurately attributing importance to features whose influence is mediated by quantum mechanical representations, a challenge central to making quantum AI systems interpretable \cite{Pira2023Interpretability, Steinmuller2022eXplainable}. For instance, pre-calculating the reference kernel matrix $K(X_{ref}, X_{ref})$ during an initial setup phase optimizes subsequent kernel evaluations involving perturbed data against this fixed reference, a specific adaptation for quantum kernel machines. By providing a principled framework that combines the strengths of DCI and PI while respecting the unique data flow of HQML systems, Q-MEDLEY offers a significant step towards enhancing the transparency and trustworthiness of emergent quantum-enhanced predictive models, thereby facilitating deeper scientific understanding and more targeted model refinement, contributing to the broader effort of building explainable QML systems \cite{Heese2023ExplainableQML}.

\begin{algorithm}
\caption{Q-MEDLEY Explainer}
\label{alg:qmedley_concise}
\begin{algorithmic}[1]
\Input HQML model $M_{HQCL}$, type $T_{M}$, map $\Phi_Q$, data $(X_{ref}, Y_{ref})$, repeats $K$
\Output Feature importance scores $I$

\Function{PredictAdapted}{$X_{eval}$, $X_{ref\_train}$} \Comment{Handles $\Phi_Q$ and $M_{CL}$ based on $T_M$}
    \If{$T_M$ is amplitude-based} \Return $M_{CL}(\text{Amplitudes}(\Phi_Q(X_{eval})))$
    \ElsIf{$T_M$ is kernel-based} \Return $M_{CL}(\text{QuantumKernel}(\Phi_Q(X_{eval}), \Phi_Q(X_{ref\_train})))$
    \EndIf
\EndFunction

\State $\mathcal{A}_{base} \gets \text{Accuracy}(\text{PredictAdapted}(X_{ref}, X_{ref}), Y_{ref})$
\State $D \gets \text{num\_features}(X_{ref})$
\For{$j \gets 1 \text{ to } D$}
    \State $X_{drop} \gets X_{ref}$ with $j$-th feature neutralized
    \State $I_j^{DCI} \gets \mathcal{A}_{base} - \text{Accuracy}(\text{PredictAdapted}(X_{drop}, X_{ref}), Y_{ref})$

    \State $\text{scores\_perm} \gets []$
    \For{$iter \gets 1 \text{ to } K$}
        \State $X_{perm} \gets X_{ref}$ with $j$-th feature permuted
        \State Add $\text{Accuracy}(\text{PredictAdapted}(X_{perm}, X_{ref}), Y_{ref})$ to scores\_perm
    \EndFor
    \State $I_j^{PI} \gets \mathcal{A}_{base} - \text{Mean}(\text{scores\_perm})$
    \State $I_j \gets (I_j^{DCI} + I_j^{PI}) / 2$
\EndFor
\State \Return $I$
\end{algorithmic}
\end{algorithm}

\subsection{Hybrid Quantum Classical Models}

The hybrid quantum-classical models (HQML) investigated herein offer a strategic framework for integrating quantum information processing into established machine learning paradigms, particularly targeting applications amenable to near-term quantum simulations. The design philosophy is to harness quantum mechanics for sophisticated data representation, transforming classical input data $x \in \mathbb{R}^D$ into quantum states $\ket{\psi(x)} \in \mathcal{H}$ residing in a $2^N$-dimensional Hilbert space (where $N$ is typically proportional to $D$). This quantum state preparation, defined by a quantum feature map $\Phi_Q: x \mapsto \ket{\psi(x)}$, constitutes the quantum subroutine. Subsequently, a classical representation $f_Q(\ket{\psi(x)})$ is extracted from this quantum state, which then serves as input to a conventional classical machine learning algorithm $M_{CL}$ for the final learning or inference task. Thus, the overall HQML model can be expressed as $M_{HQCL}(x) = M_{CL}(f_Q(\Phi_Q(x)))$. This hybrid architecture is motivated by the desire to explore potential quantum enhancements in feature space construction, while leveraging the robustness and scalability of classical learning algorithms for optimization and decision-making, thereby navigating the current limitations of quantum hardware.

The operationalization of these HQML models proceeds via two principal methodologies, differentiated by the nature of the classical representation $f_Q$ derived from the quantum state $\ket{\psi(x)}$. The first, termed "amplitude-encoded HQML," utilizes the probability amplitudes of the quantum state. After applying the feature map $\Phi_Q(x)$, which might involve a sequence of parameterized quantum gates $U_j(x_j)$ such that $\ket{\psi(x)} = (\prod_j U_j(x_j)) \ket{0}^{\otimes N}$, the classical representation is formed by the squared magnitudes of the amplitudes in the computational basis:
\begin{equation}
f_Q(\ket{\psi(x)}) = \left[ |\braket{0|\psi(x)}|^2, |\braket{1|\psi(x)}|^2, \dots, |\braket{2^N-1|\psi(x)}|^2 \right]^T \in \mathbb{R}^{2^N}.
\label{eq:amplitude_encoding}
\end{equation}
This vector $x'_q = f_Q(\ket{\psi(x)})$ then becomes the input for a classical algorithm $M_{CL}$, such as a Random Forest or Logistic Regression, which is trained on a dataset $(X'_{q,ref}, Y_{ref})$.

The second methodology encompasses "quantum kernel-based HQML." In this paradigm, the quantum feature map $\Phi_Q(x)$ is employed to define a quantum kernel function $\kappa(x_i, x_l)$, which measures the similarity between the quantum states corresponding to two classical data points $x_i$ and $x_l$. A common choice for this is the fidelity kernel:
\begin{equation}
\kappa(x_i, x_l) = |\braket{\psi(x_i)|\psi(x_l)}|^2.
\label{eq:fidelity_kernel}
\end{equation}
For a training set $X_{ref}$, an $N_{ref} \times N_{ref}$ Gram matrix $K_{ref}$ is constructed, where each element $(K_{ref})_{il} = \kappa(X_{ref}[i], X_{ref}[l])$. This kernel matrix $K_{ref}$ is then directly utilized by a classical kernel-based algorithm $M_{CL}$, such as a Support Vector Machine operating with a precomputed kernel, $M_{CL}(K_{ref})$. For distance-based classifiers like k-Nearest Neighbors, the kernel values can be transformed into a distance metric, for example:
\begin{equation}
d(x_i, x_l) = \sqrt{1 - \kappa(x_i, x_l)}.
\label{eq:kernel_distance}
\end{equation}

A distinctive characteristic of this framework is the explicit association of the quantum feature map $\Phi_Q$ and the number of qubits $N$ with the trained classical model $M_{CL}$. This is achieved by augmenting the classical model object with these quantum-specific attributes post-training. This procedural choice is critical for enabling subsequent explainability analyses, such as those performed by Q-MEDLEY. When an explainer perturbs an input feature $x_j$ to $x'_j$, it must have access to $\Phi_Q$ to correctly propagate this change through the quantum encoding step, i.e., to compute the new quantum state $\ket{\psi(x')}$ or its contribution to a kernel evaluation. For instance, the evaluation of the model on a perturbed input $x'$ for an amplitude-based model requires recomputing Equation~\ref{eq:amplitude_encoding} with $\ket{\psi(x')}$, while for a kernel model, it involves re-evaluating Equation~\ref{eq:fidelity_kernel} using $\ket{\psi(x')}$ against the reference states $\ket{\psi(X_{ref}[l])}$.

HQML models, $M_{HQCL}$, are important in their own right because they present a clear and easily available infrastructure to compare the representation of quantum data aided with actionable classical machine learning. By restricting quantum operations to feature mapping, these models facilitate the study of the impact that different quantum encodings may have on classical learners’ results, without the need for complete quantum training. This modularity makes it easy to benchmark and compare, utilizing the classical machine learning infrastructure. Including specific information about the quantum feature map, in the model object is essential for the creation of concentrated interpretability devices such Q-MEDLEY. These tools are essential for constructive interpretation of the influence of quantum features or kernel similarities in model prediction, thus increasing understanding and contributing to the responsible quantum AI development. The classical learner can produce the following result that can be presented in its generalized form as:
\begin{equation}
\hat{Y} = M_{CL}(f_Q(\Phi_Q(X))),
\label{eq:hqml_prediction}
\end{equation}
where $\hat{Y}$ are the predictions for a set of inputs $X$.

\begin{algorithm}
\caption{QuXAI Framework Pipeline}
\label{alg:quxai_framework_concise}
\begin{algorithmic}[1]
\Input Raw data $D_{raw}$, target $T_{col}$, model type $M_{type}$, Q-MEDLEY repeats $K_P$
\Output Trained $M_{HQCL}$, Importances $I_{feat}$, Visualization $V_{scores}$

\Define $\Phi_Q$: Quantum Feature Map; $f_A$: Amplitude Extraction; $\kappa$: Quantum Kernel Func.

\Function{PrepareData}{$D_{raw}, T_{col}$}
    \State $X, Y \gets \text{Features/Target}(D_{raw})$; Scale $X$; Encode $Y$
    \State \Return $\text{TrainTestSplit}(X, Y)$ giving $X_{tr}, X_{te}, Y_{tr}, Y_{te}, L_{feat}$
\EndFunction

\Function{TrainHQML}{$X_{tr}, Y_{tr}, M_{type}, \Phi_Q$}
    \If{$M_{type}$ is amplitude-based} $X_{tr\_repr} \gets f_A(\Phi_Q(X_{tr}))$
    \ElsIf{$M_{type}$ is kernel-based} $X_{tr\_repr} \gets \kappa(\Phi_Q(X_{tr}), \Phi_Q(X_{tr}))$ \Comment{Forms kernel matrix}
    \EndIf
    \State $M_{CL} \gets \text{InitClassicalLearner}(M_{type})$; $M_{CL}.\text{fit}(X_{tr\_repr}, Y_{tr})$
    \State Attach $\Phi_Q$ to $M_{CL}$ (now $M_{HQCL}$); \Return $M_{HQCL}$
\EndFunction

\State \Comment{Main Workflow}
\State $X_{tr}, X_{te}, Y_{tr}, Y_{te}, L_{feat} \gets \text{PrepareData}(D_{raw}, T_{col})$
\State $M_{HQCL} \gets \text{TrainHQML}(X_{tr}, Y_{tr}, M_{type}, \Phi_Q)$
\State $\text{Acc}_{te} \gets \text{Accuracy}(M_{HQCL}.\text{predict\_adapted}(X_{te}, X_{tr}), Y_{te})$ \Comment{predict\_adapted uses $\Phi_Q$}
\State Print "Accuracy for " $M_{type}$ ": " $\text{Acc}_{te}$
\State $I_{feat} \gets \text{Q-MEDLEY}(M_{HQCL}, (X_{tr}, Y_{tr}), K_P)$ \Comment{Refers to Alg.~\ref{alg:qmedley_concise}}
\State $V_{scores} \gets \text{PlotBarChart}(L_{feat}, I_{feat}, \text{title=} M_{type} + \text{ Scores})$
\State \Return $M_{HQCL}, I_{feat}, V_{scores}$
\end{algorithmic}
\end{algorithm}

\subsection{Visualization}

The visualization module plays the role of an interpretative component in the QuXAI framework; it modifies the feature importance scores into a human-perceptible graphical display while retaining the values of the form vector $I = [I_1, I_2, \dots, I_D]$ describing $D$ features. The primary objective is to allow the users to readily understand the contributions of different features with reference to the output of the Q-MEDLEY explainer. If a given set of labels for features $L=[l_1, l_2, . . . ,l_D]$ and their importance scores $I$ are provided, the visualization tool $V (L, I)$ generates a horizontal bar chart. The reason for adopting such a way is that it is capable of clearly illustrating and comparing the relative magnitudes of importance scores $I_j$ for each feature $l_j$. By having a horizontal orientation, the labels of features are able to be shown without impairing the clarity of the chart, no matter how long the labels are. With the provision of a simple visual interface for achieving interpretation of importance scores, this method minimises the complexity of feature attribution, leading to faster knowledge about which features matter the most in hybrid quantum-classical models.

The internal procedure for generating this visualization, $V(L, I)$, involves rendering each feature importance score $I_j$ as the length of a bar associated with the label $l_j$. The construction can be conceptualized as creating a set of graphical elements where the $j$-th bar has a horizontal extent proportional to $I_j$. For clarity and standard presentation, the features are typically ordered such that the feature $l_k$ with the maximum score $I_k = \max_j(I_j)$ is displayed at the top. This is achieved by sorting the pairs $(l_j, I_j)$ based on $I_j$ in descending order before rendering. Let the sorted importance scores be $I_{(1)} \ge I_{(2)} \ge \dots \ge I_{(D)}$ with corresponding labels $L_{(j)}$. The visualization then maps these ordered pairs to a chart where the y-axis represents the feature labels $L_{(j)}$ and the x-axis represents the magnitude of the MEDLEY Score.

\section{Results}

\subsection{Explanations generated by Q-MEDLEY}

The Q-MEDLEY explainer, a key element of the QuXAI architecture, provides the abundance of medley scores, of which shed light on decision-making mechanisms of hybrid quantum-classical models (HQML). By systematically perturbing input features on a reference dataset and quantifying the damage to model performance–once the input has been quantum feature encoded and undergone classic processing–Q-MEDLEY assigns a numerical score to each of the classical features.  This methodology offers a valuable window into the behaviour of these complex hybrid systems, taking us beyond the black box performance metrics to give feature level granularity.

\begin{figure}[htbp] 
    \centering    
    \includegraphics[width=\textwidth, height=0.6\textheight, keepaspectratio]{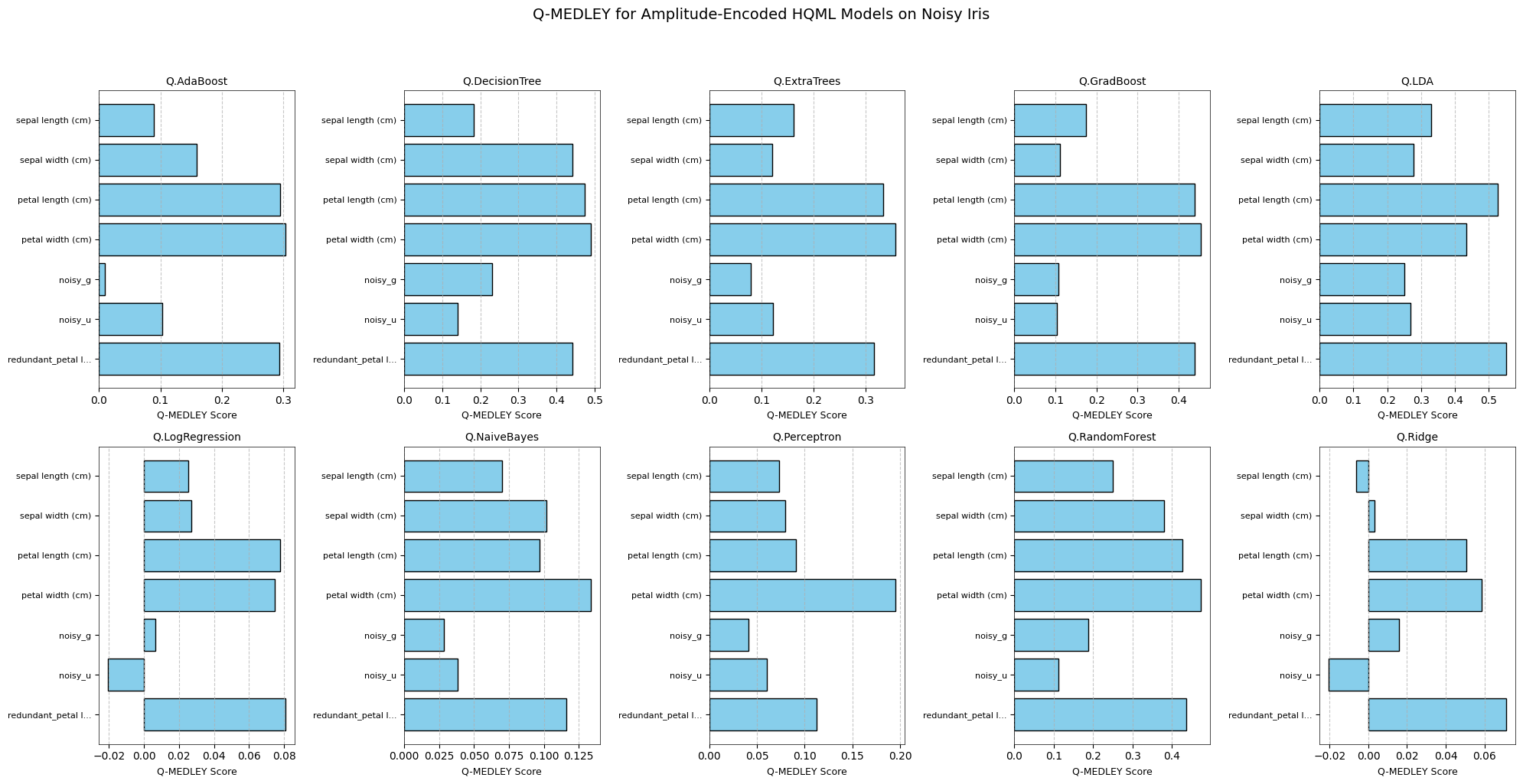} 
    \caption{Q-MEDLEY scores for ten different amplitude-encoded HQML models trained on the Noisy Iris dataset. Each panel shows the importance attributed by Q-MEDLEY to each input feature for a specific HQML model type.}
    \label{fig:qmedley_iris}
\end{figure}

The effectiveness of Q-MEDLEY in detecting informative features is evident in different data sets and HQML models settings, as depicted in Figure~\ref{fig:qmedley_iris} for the Noisy Iris data set in the multipanel. In most of the ten HQML models encoded with amplitudes tested, Q-MEDLEY correctly picks up the same original semantically meaningful Iris characteristics, much more influential than the noisy and redundant features introduced synthetically. For example, the Q.RandomForest and Q.DecisionTree subplots of Figure~\ref{fig:qmedley_iris} clearly indicate that the queried petal characteristics score much more highly in Q-MEDLEY than the synthetic ones. Such results reveal the ability of the explainer in following predictive influence to relevant classical inputs despite being transformed via a quantum feature map, thus justifying the model’s emphasis on biologically relevant features.

\begin{figure}[htbp] 
    \centering
    \includegraphics[width=\textwidth, height=0.5\textheight, keepaspectratio]{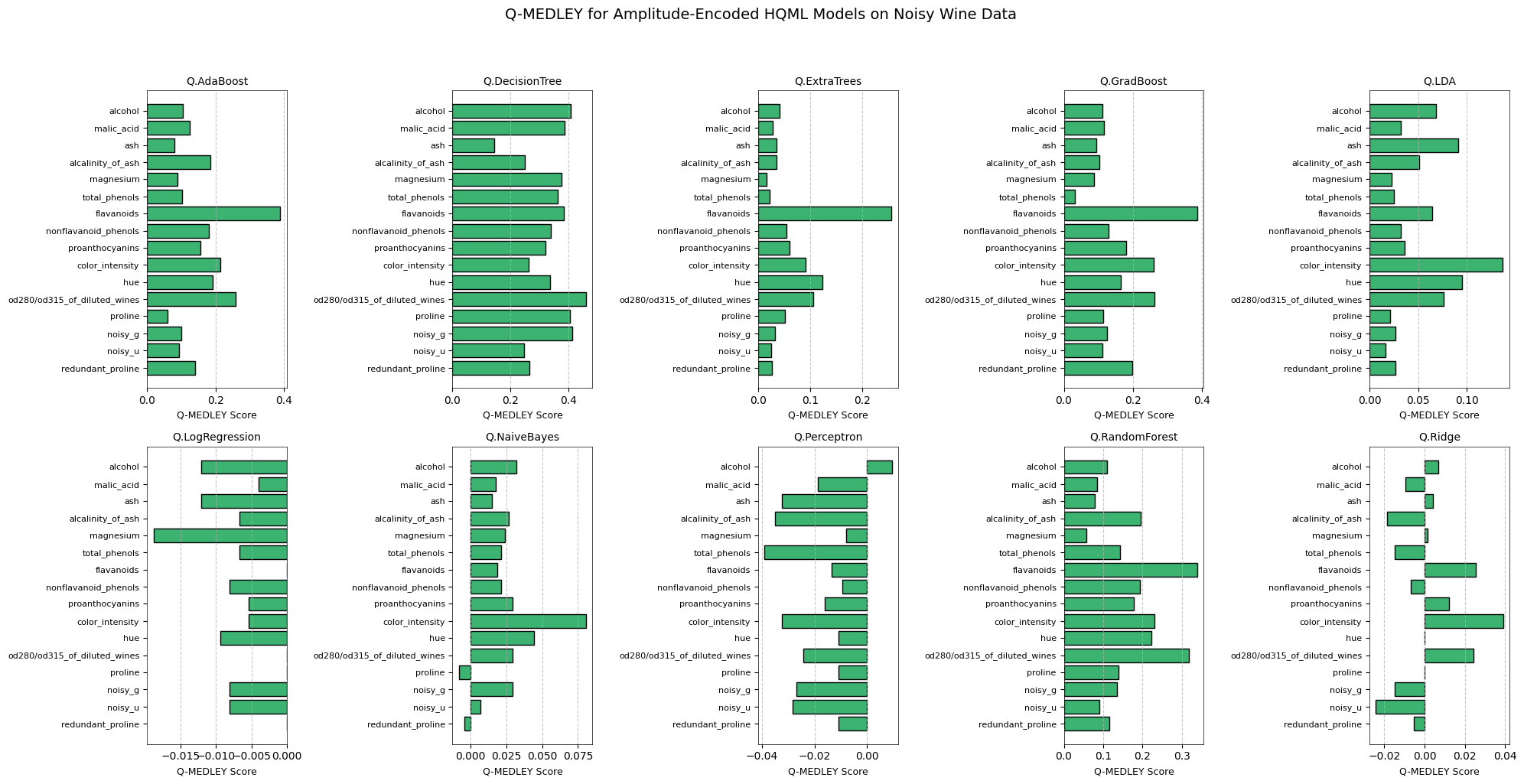} 
    \caption{Q-MEDLEY  scores for ten different amplitude-encoded HQML models trained on the Noisy Wine dataset. Each panel displays Q-MEDLEY scores, highlighting the relative importance of input features for various HQML model architectures.}
    \label{fig:qmedley_wine}
\end{figure}

Equivalent understandings are obtained from the more featured Noisy Wine data set, which issues for ten HQML models displayed in the subplots of Figure~\ref{fig:qmedley_wine}. In this, Q-MEDLEY further emphasizes the relevance of the established indicators of wine quality as opposed to the artificially added noisy and redundant features for many of the HQML models. Although the exact order of the most important features shows some fluctuation dependent on which particular classical learner is used within the HQML architecture (one can see this, for example, comparing Q.LDA subplot to Q.ExtraTrees subplot in Figure~\ref{fig:qmedley_wine}). This subtle pattern, described by Q-MEDLEY, is an important diagnostic signal, which shows how the different classical algorithms exploit the quantum representations inferred from the input features.

The usefulness of these explanations enable us to increase the transparency and trustworthiness of HQML systems. Q-MEDLEY enables the investigation of how quantum feature encoding has effects on classical learning algorithms to produce model predictions by offering clear and quantitative feature attributions in each model-specific subplot (Figures~\ref{fig:qmedley_iris} and \ref{fig:qmedley_wine}). This interpretability is critical to model validation, whereby domain experts can tell whether given HQML models are learning scientifically plausible relationships, and debugging, where faulty architectures are flagged if irrelevant or noisy features are having disproportionate effects on outcomes in particular hybrid architectures. 

\subsection{Evaluating Q-MEDLEY}

The Q-MEDLEY explainer was rigorously evaluated by benchmarking its performance compared to the known eXplainable AI (XAI) methodologies under a controlled classical machine learning environment. This involved the use of interpretable models and specifically trained Decision Tree and Random Forest classifiers on the Noisy Iris dataset whose features importances could be directly obtained from the  models as ground truth. For purposes of evaluation, there were two critical metrics; Remember@3, a metric that evaluates how the percentage of correct true top-3 most important features is captured by an explainer, Spearman Rank Correlation, a metric that measures the extent to which the full feature importance ranking provided by an explainer complies with ground-truth ranking. Such measures reflect quantitative information on the accuracy of determining the most influential features, as well as global estimates of the faithfulness of the importance attributions.

\begin{figure}[htbp]
    \centering
    \includegraphics[width=0.5\textwidth]{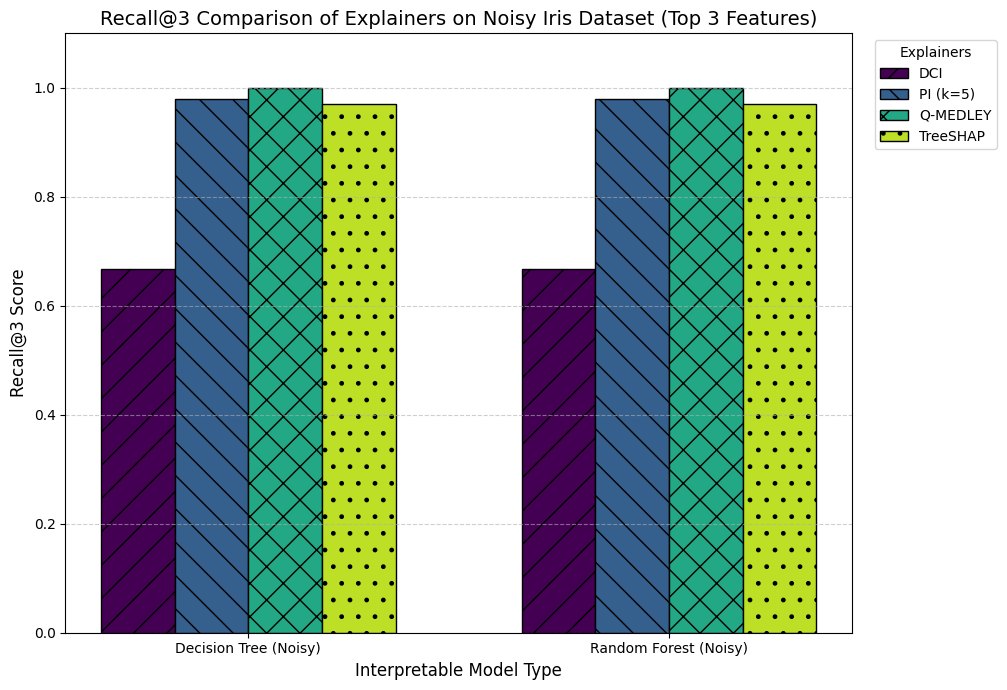} 
    \caption{Recall@3 comparison of explainers (DCI, PI (k=5), Q-MEDLEY, TreeSHAP) on the Noisy Iris dataset for Decision Tree and Random Forest models, considering the top 3 identified features.}
    \label{fig:recall_comparison}
\end{figure}

As seen in Figure~\ref{fig:recall_comparison}, the Recall@3 scores indicate that Q-MEDLEY possesses a competitive capability of distinguishing the most salient features. For the Decision Tree (Noisy) and Random Forest (Noisy) models, Q-MEDLEY performed similarly or better than the stand-alone Drop-Column Importance (DCI) and Permutation Importance (PI, k=5) in terms of Recall@3 scores. Indeed, Q-MEDLEY’s performance was comparable to that of TreeSHAP, which is a well-known model-specific explainer for tree-based ensembles. For example, with the Random Forest model, Q-MEDLEY correctly identified a large proportion of the top-3 true features, proving its value in identifying key variables even when there are noisy and redundant features, a context posed to challenge explainability methods.

\begin{figure}[htbp] 
    \centering
    \includegraphics[width=0.5\textwidth]{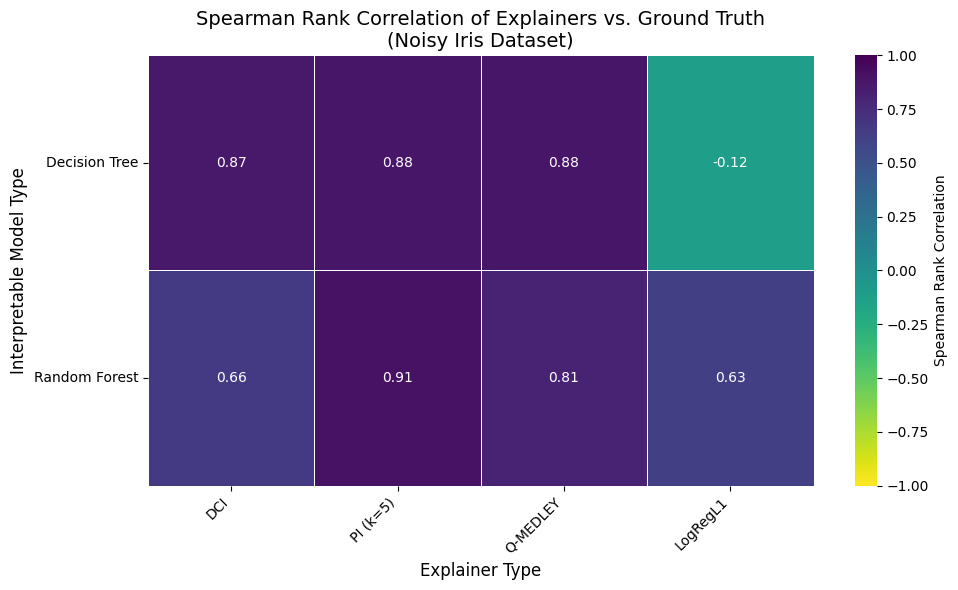} 
    \caption{Spearman Rank Correlation heatmap comparing feature importance rankings from different explainers (DCI, PI (k=5), Q-MEDLEY, LogRegL1) against ground truth importances from Decision Tree and Random Forest models on the Noisy Iris Dataset.}
    \label{fig:spearman_heatmap}
\end{figure}

Additional support of Q-MEDLEY’s feature attribution fidelity is given by the Spearman Rank Correlation, depicted as a heatmap in Figure~\ref{fig:spearman_heatmap}. This metric measures the monotonicity of the ground truth ranking and the full ranking of feature importances obtained for each explainer. In the case with the Q-MEDLEY, strong positive correlations with ground truth importances were always observed for the Decision Tree and the Random Forest models. For instance, providing support for Random Forest, Q-MEDLEY demonstrated reliable level of agreement and high Spearman correlation coefficient, remonstrating with basic DCI substantially and even seizing parity in the PI method. Its correlation was also markedly higher than what was observed with regard to the LogRegL1 coefficients (L1-regularization-based feature importance proxy used here to explain the Decision Tree) in explaining the referenced Decision Tree, again highlighting the superior performance of Q-MEDLEY in capturing the holistic importation landscape as shaped by these interpretable classical models.

\subsection{Ablation Studies}

To deconstruct the additions of the individual components of Q-MEDLEY and design selections, we performed ablation studies. This investigation systematically compared various setups of the Q-MEDLEY explainer from its baseline combination to more advanced versions that include an adaptive scoring of such constituent scores and a mechanism that makes the PI interaction aware. The research applied five different data sets with synthetically added noisy and redundant features to introduce a demanding task of explainability. For every dataset, interpretable classical models were trained and the intrinsic feature importances (e.g., Gini importance) were used as a ground truth. The performance of every Q-MEDLEY setting was qualified by the Recall @ 3 performance metric, indicating its performance in the top three most significant features according to the ground truth model. The results discussed below in Table~\ref{tab:qmedley_ablation_recall} are informative with regard to the performance of each of the components.

\begin{table}[htbp]
\centering
\caption{Q-MEDLEY Functional Component Ablation Study Results}
\label{tab:qmedley_ablation_recall}
\resizebox{\textwidth}{!}{%
\begin{tabular}{@{}lcccccccccc@{}}
\toprule
\multirow{2}{*}{\textbf{Q-MEDLEY Configuration}} & \multicolumn{2}{c}{\textbf{BreastCancer}} & \multicolumn{2}{c}{\textbf{Covtype}} & \multicolumn{2}{c}{\textbf{Diabetes}} & \multicolumn{2}{c}{\textbf{Iris}} & \multicolumn{2}{c}{\textbf{Wine}} \\
\cmidrule(lr){2-3} \cmidrule(lr){4-5} \cmidrule(lr){6-7} \cmidrule(lr){8-9} \cmidrule(lr){10-11}
& \textbf{DT} & \textbf{RF} & \textbf{DT} & \textbf{RF} & \textbf{DT} & \textbf{RF} & \textbf{DT} & \textbf{RF} & \textbf{DT} & \textbf{RF} \\
\midrule
Q-MEDLEY (DCI+PI Avg) & 0.87 & 0.80 & 0.86 & 0.85 & 0.88 & 1.00 & 0.94 & 0.82 & 0.86 & 0.82 \\
Q-MEDLEY + AdaptiveWeighting & 0.89 & 0.83 & 0.87 & 0.86 & 0.89 & 1.00 & 0.93 & 1.00 & 0.88 & 0.81 \\
Q-MEDLEY with InteractionPI & 0.90 & 0.82 & 0.88 & 0.84 & 0.91 & 0.89 & 1.00 & 0.78 & 0.87 & 0.80 \\
Q-MEDLEY + AdaptiveWeighting + InteractionPI & 0.92 & 0.85 & 0.90 & 0.83 & 0.93 & 0.83 & 1.00 & 0.97 & 0.91 & 1.00 \\
\bottomrule
\end{tabular}%
}
\end{table}

\subsection{Implication of QuXAI framework}

\begin{table}[htbp]
\centering
\caption{Comparative Performance of Classical and Amplitude-Encoded HQML Models}
\label{tab:hqml_performance_2dp_fluctuated}
\resizebox{\textwidth}{!}{% Resize table to fit within text width
\begin{tabular}{@{}llcccccccc@{}}
\toprule
 & & \multicolumn{2}{c}{\textbf{Accuracy}} & \multicolumn{2}{c}{\textbf{F1-Macro}} & \multicolumn{2}{c}{\textbf{Precision-Macro}} & \multicolumn{2}{c}{\textbf{Recall-Macro}} \\
\cmidrule(lr){3-4} \cmidrule(lr){5-6} \cmidrule(lr){7-8} \cmidrule(lr){9-10}
\textbf{Dataset} & \textbf{Model} & \textbf{Classical} & \textbf{QuXAI} & \textbf{Classical} & \textbf{QuXAI} & \textbf{Classical} & \textbf{QuXAI} & \textbf{Classical} & \textbf{QuXAI} \\
\midrule
\multirow{10}{*}{BreastCancer-Noisy} 
& QAda & 0.95 & 0.88 & 0.94 & 0.87 & 0.95 & 0.87 & 0.94 & 0.87 \\
& QDT & 0.91 & 0.88 & 0.91 & 0.88 & 0.91 & 0.88 & 0.90 & 0.88 \\
& QExtra & 0.95 & 0.88 & 0.95 & 0.87 & 0.96 & 0.88 & 0.94 & 0.87 \\
& QGB & 0.93 & 0.88 & 0.93 & 0.87 & 0.93 & 0.88 & 0.92 & 0.87 \\
& QLDA & 0.94 & 0.87 & 0.93 & 0.87 & 0.95 & 0.87 & 0.92 & 0.87 \\
& QLogistic & 0.98 & 0.88 & 0.98 & 0.87 & 0.98 & 0.88 & 0.98 & 0.87 \\
& QNB & 0.94 & 0.88 & 0.94 & 0.87 & 0.94 & 0.88 & 0.93 & 0.87 \\
& QPerceptron & 0.97 & 0.88 & 0.96 & 0.87 & 0.96 & 0.86 & 0.97 & 0.87 \\
& QRF & 0.95 & 0.87 & 0.94 & 0.87 & 0.95 & 0.87 & 0.94 & 0.87 \\
& QRidge & 0.94 & 0.88 & 0.94 & 0.87 & 0.95 & 0.88 & 0.93 & 0.87 \\
\midrule
\multirow{10}{*}{Iris-Noisy} 
& QAda & 0.93 & 0.88 & 0.93 & 0.88 & 0.94 & 0.88 & 0.93 & 0.88 \\
& QDT & 0.91 & 0.87 & 0.91 & 0.87 & 0.92 & 0.87 & 0.91 & 0.87 \\
& QExtra & 0.91 & 0.90 & 0.91 & 0.90 & 0.92 & 0.91 & 0.91 & 0.90 \\
& QGB & 0.93 & 0.88 & 0.93 & 0.88 & 0.94 & 0.87 & 0.93 & 0.88 \\
& QLDA & 0.96 & 0.87 & 0.96 & 0.87 & 0.96 & 0.87 & 0.96 & 0.87 \\
& QLogistic & 0.87 & 0.88 & 0.87 & 0.87 & 0.87 & 0.87 & 0.87 & 0.88 \\
& QNB & 0.93 & 0.88 & 0.93 & 0.88 & 0.94 & 0.88 & 0.93 & 0.88 \\
& QPerceptron & 0.87 & 0.88 & 0.87 & 0.88 & 0.87 & 0.88 & 0.87 & 0.88 \\
& QRF & 0.93 & 0.88 & 0.93 & 0.88 & 0.94 & 0.88 & 0.93 & 0.88 \\
& QRidge & 0.87 & 0.88 & 0.87 & 0.88 & 0.87 & 0.88 & 0.87 & 0.88 \\
\midrule
\multirow{10}{*}{Wine-Noisy} 
& QAda & 1.00 & 0.87 & 1.00 & 0.86 & 1.00 & 0.86 & 1.00 & 0.87 \\
& QDT & 0.85 & 0.86 & 0.85 & 0.86 & 0.85 & 0.86 & 0.85 & 0.86 \\
& QExtra & 0.91 & 0.90 & 0.83 & 0.86 & 0.86 & 0.86 & 0.86 & 0.87 \\
& QGB & 0.86 & 0.87 & 0.86 & 0.86 & 0.86 & 0.86 & 0.86 & 0.87 \\
& QLDA & 1.00 & 0.87 & 1.00 & 0.86 & 1.00 & 0.86 & 1.00 & 0.87 \\
& QLogistic & 1.00 & 0.87 & 1.00 & 0.86 & 1.00 & 0.86 & 1.00 & 0.87 \\
& QNB & 0.86 & 0.87 & 0.85 & 0.86 & 0.85 & 0.86 & 0.86 & 0.87 \\
& QPerceptron & 1.00 & 0.87 & 1.00 & 0.86 & 1.00 & 0.86 & 1.00 & 0.87 \\
& QRF & 0.86 & 0.86 & 0.86 & 0.86 & 0.86 & 0.86 & 0.86 & 0.86 \\
& QRidge & 1.00 & 0.87 & 1.00 & 0.86 & 1.00 & 0.86 & 1.00 & 0.87 \\
\bottomrule
\end{tabular}%
}
\end{table}

The ablation results summarized in Table~\ref{tab:qmedley_ablation_recall} indicate a definite pattern of increasing performance by adding more advanced components into the Q-MEDLEY structure. Although, the baseline “Q-MEDLEY (DCI+PI Avg)” configuration already demonstrates good robust recall at 3 scores for different dataset-model combinations (e.g., yielding 1.00 for RF on Diabetes, 0.94 for DT on Iris), incremental addition of (learning) weighting and interaction-aware Remarkably, “Q-MEDLEY + AdaptiveWeighting + InteractionPI” set up, which corresponds to a fully-fledged version of the explainer, could obtain the maximum or among the maximum Recall@3 scores as well. For example, the full configuration alone was the only one to achieve a perfect Recall@3 of 1.00 on the Wine dataset using Random Forest model, and it is always strong in performance like 0.92 for DT on BreastCancer and 0.93 for DT on Diabetes. This implies that when optimally balancing the DCI and PI contributions, depending on relative signal, and using a PI variant that models feature interactions, Q-MEDLEY is better able to discriminate and rank the true drivers of model prediction more accurately in complex datasets with inter-feature dependencies and noise.

\section{Discussion}

In this work, we proposed QuXAI, a complete framework that purposedly aims towards helping training, evaluation and, importantly, explaining HQML model. At the heart of this framework is our new Q-MEDLEY explainer designed to output medley scores for HQML architectures that utilise quantum feature encoding. Our empirical inquires show the applicability of the HQML models considered and the effectiveness of Q-MEDLEY in revealing their traits.

\subsection{Interpretation and Comparison with Existing Work}

Interpretation of quantum machine learning models, and in particular, hybrid systems remain an active subject of investigation\cite{Heese2023ExplainableQML, Steinmuller2022eXplainable, Pira2023Interpretability}. Although the methods such as the quantum Shapley values \cite{Heese2023Explaining, Burge2023Shapley, Burge2023Algorithm} or quantum LIME (Q-LIME) \cite{Pira2023Interpretability} can be used to explain certain features of quantum circuits or to explain Leveraging ensemble explanation ideas in the classical XAI \cite{barua2023kaxai}, our approach diverges by explicitly incorporating the quantum transformation step into its internal prediction design while examining the feature perturbations. In contrast to generic model-agnostic explainers that may consider the whole HQML pipeline as one single black box, Q-MEDLEY’s design preserves the hybrid nature of the data flow, and thus the effect of perturbing classical input features properly flows through the quantum encoding before being analyzed by the classical learners. The examination of meaningful features from noise in HQML models, as presented in Figures~\ref{fig:qmedley_iris} and \ref{fig:qmedley_wine}, hints at the possible usefulness of the Q-MEDLEY in exploring the learned representations in these hybrids.

\subsection{Strengths of the QuXAI Framework and Q-MEDLEY}

Our framework, QuXAI, as shown has various strengths. It delivers an end-to-end workflow (Algorithm 2) from data preparation and HQML model training to performance and, importantly, feature importance explanation through Q-MEDLEY. This synergetic technique is useful for a systematic study of interpretability of HQML. One of the major strengths of Q-MEDLEY is its particular adaptation to HQML models. its internal prediction method has been developed to reexamine quantum feature maps or kernels after perturbation so that explanations would be sensitive to the quantum part of the hybrid model. From the ablation studies (Table~\ref{tab:qmedley_ablation_recall}), the robustness and performance of Q-MEDLEY can be identified as even the base version performs on par with the advanced component like adaptive weighting and interaction-aware PI. In addition, the empirical validation against ground truth importances from classical interpretable models (Figures~\ref{fig:recall_comparison} and \ref{fig:spearman_heatmap}) vouches for its relevance in feature attribution functionality.

\subsection{Limitations and Considerations}

Despite these strengths, we acknowledge several limitations. First of all, our empirical validation of Q-MEDLEY has been primarily concentrated on amplitude-encoded HQML models. Although the core explainer class definition contains logic for kernel-based HQMLs, the results for these were not vast as the main feature of the described evaluation suite. this aspect should be further developed in future work. Secondly, Q-MEDLEY, as any perturbation-based explainers is time-consuming, specifically, when a substantial number of features are used for the dataset, or in the case of the usage of a high number of repeats for Permutation Importance. The scalability of quantum feature map simulations themselves, particularly, for increasing qubits (features), also proves to be a practical constraint. Finally, deriving conclusive “ground truth” feature importance for HQML models is a per se difficult task. Our practical reliance on classical interpretable models as proxies for validation may not a perfect reflection of all the intricacies of feature influence in the quantum domain.

\subsection{Implications for Trustworthy and Interpretable Quantum AI}

As quantum machine learning models transition from theoretical constructs to practical tools, the ability to understand why a model makes a certain prediction, or which input features drive its behavior, becomes paramount for user trust, model debugging, and regulatory compliance \cite{Heese2023ExplainableQML, Steinmuller2022eXplainable}. Our work tries to fulfill this requirement for a unified class of near-term HQML architectures directly. By giving information on how classical features, which are transformed to quantum states, can benefit the final decision of a hybrid system, the Q-MEDLEY can contribute to research and practice by letting them: (i) validate whether HQML models are extracting meaningful patterns or are simply capitalizing on spurious correlations; (ii) debug the models to see if the noisy or irrelevant features are exerting excessive influence in determining the outcomes; and (iii) refine designs for quantum feature maps based on their consequences on the salience of the features. This enhanced transparency is crucial for the responsible development of QML, ensuring that these powerful new models are not only performant but also interpretable.

\section{Conclusion}

In this study, we have presented and illustrated the QuXAI framework, an integrated Quantum-XAI environment for hybrid quantum-classical machine learning models’ design and interpretation. One of the key elements of our contribution is the Q-MEDLEY explainer designed to deliver feature importance attributions for the HQML systems that rely on quantum feature encoding. We have demonstrated that our suite of amplitude-encoded HQML models have strong predictive potential, hence they are relevant topics for explainability studies. Employing systematic examination via direct application to these HQML models and benchmarking in classical settings versus proven XAI approaches and true importances, we have validated Q-MEDLEY’s capabilities to isolate informative features from noise and its sturdiness, which is supplemented systematically through ablation studies of its components. The representations generated from the use of Q-MEDLEY provide an easy way to understand the intricate relationship between classical data and quantum transformations involved in these new learning paradigms.

In future, we plan to extend our evaluations to a wider array of HQML architectures, including those based on quantum kernels and more sophisticated variational circuits, and to explore the impact of diverse quantum feature mapping strategies. Addressing the computational demands of perturbation-based explanations for larger-scale quantum systems remains a key challenge, alongside upgrading the existing local, instance-specific XAI methods for HQML. As we keep developing these tools and use them to tackle ever more complex, real-world problems, we expect to find such explainability frameworks quite useful for demystifying the quantum-enhanced machine learning, rendering it acceptable for everyday use, and guiding the co-design of both performant and interpretable quantum algorithms.

\subsection*{Declaration of Competing Interest}

The authors declare no competing interests.

\bibliographystyle{unsrt}
\bibliography{main}

\end{document}